\documentclass[english]{lni}

\usepackage{graphicx}
\usepackage{fancyhdr}
\usepackage{changepage} 
\usepackage{listings} 
\usepackage{xcolor}
\usepackage{algorithm}
\usepackage{amsmath}
\usepackage{amssymb}
\usepackage{booktabs}
\usepackage{color}
\usepackage{url}
\usepackage{arydshln}
\usepackage[noend]{algpseudocode}

\usepackage[figurename=Fig., tablename=Tab., small]{caption}

\fancypagestyle{titlepage}{
\fancyhead[RO]{\small F.  Boutros,  N. Damer,  M. Fang,  M. Gomez-Barrero,  K. Raja,  \linebreak C. Rathgeb,   A.  Sequeira,  and M. Todisco (Eds.): BIOSIG 2024, \linebreak Lecture Notes in Informatics (LNI), Gesellschaft f\"ur Informatik, Bonn 2024} 
\fancyfoot{}}

\renewcommand{\headrulewidth}{0.4pt} 
\author{Yoshihiko Furuhashi\footnote{National Institute of Informatics, 2-1-2 Hitotsubashi, Tokyo 101-8430, Japan, \{yfuruhashi, jyamagis, wangxin, nhhuy, iechizen\}@nii.ac.jp}, 
\ Junichi Yamagishi\footnotemark[1], 
\ Xin Wang\footnotemark[1], 
\ Huy H. Nguyen\footnotemark[1], 
and Isao Echizen\footnotemark[1]}

\title{Exploring Active Data Selection Strategies for \\ Continuous Training in Deepfake Detection}
\begin{document}

\maketitle

\renewcommand{\refname}{References}
\setcounter{footnote}{2} 
\thispagestyle{titlepage}
\pagestyle{fancy}
\fancyhead{} 
\fancyhead[RO]{\small Exploring Active Data Selection Strategies for Continuous Training in Deepfake Detection \hspace{25pt}  \hspace{0.05cm}}
\fancyhead[LE]{\hspace{0.05cm}\small  \hspace{25pt} Yoshihiko Furuhashi, Junichi Yamagishi, Xin Wang, Huy H. Nguyen, and Isao Echizen}
\fancyfoot{} 
\renewcommand{\headrulewidth}{0.4pt} 

\begin{abstract}
In deepfake detection, it is essential to maintain high performance by adjusting the parameters of the detector as new deepfake methods emerge. In this paper, we propose a method to automatically and actively select the small amount of additional data required for the continuous training of deepfake detection models in situations where deepfake detection models are regularly updated. The proposed method automatically selects new training data from a \textit{redundant} pool set containing a large number of images generated by new deepfake methods and real images, using the confidence score of the deepfake detection model as a metric. Experimental results show that the deepfake detection model, continuously trained with a small amount of additional data automatically selected and added to the original training set, significantly and efficiently improved the detection performance, achieving an EER of 2.5\% with only 15\% of the amount of data in the pool set.
\end{abstract}

\section{Introduction}

Deepfake detection requires the capability to detect images and videos generated or manipulated by using unknown spoofing methods. However, this is known to be a challenging task, and research is thus actively ongoing. Various approaches have been proposed to enhance the generalization performance of deepfake detectors. These include the use of adapters \cite{liu2024forgery}, contrastive learning techniques \cite{dong2023contrastive, larue2023seeable}, the self-supervised auxiliary task \cite{das2024limited}, and curriculum learning~\cite{song2024towards}, where models are trained step by step according to image quality. Despite these advancements, deepfake detection remains an open and challenging problem \cite{yan2024deepfakebench}. 

We posit that the robust and reliable detection of unknown spoofing methods necessitates \textit{continuous training} \cite{ma2021continual} in practice. More specifically, we believe that a new continuous training framework that differs from domain adaptation is needed, which adapts existing models to new spoofing data. However, this is difficult because the ability to detect previously recognizable spoofing methods is declining or has been lost. New types of continuous training need to be explored to accommodate the detection of new spoofing methods while retaining previous detection capabilities.

\begin{figure}[t]
\centering
\includegraphics[width=0.7\columnwidth]{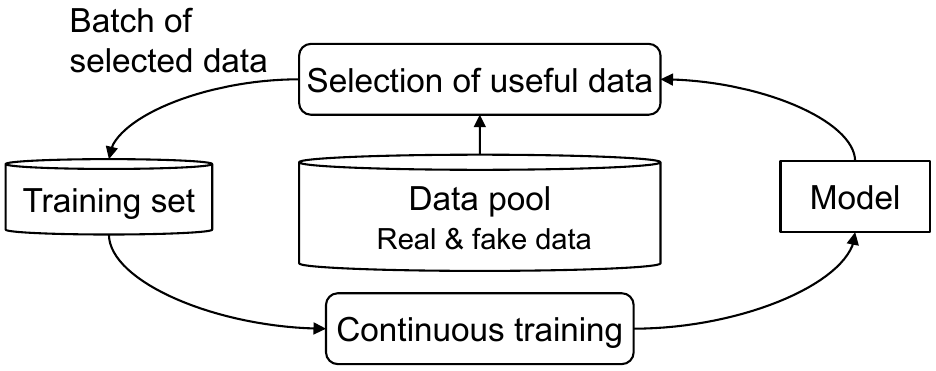}
\caption{Active data selection for continuous training of deepfake detection models.}  
\label{fig:idea}
\end{figure}

With these objectives in mind, we have developed a unique framework that diverges from both generalization and domain adaptation. We propose to automatically and actively select the small amount of additional data required for continuous training of the deepfake detection model in situations where the model is regularly updated. Figure~\ref{fig:idea} illustrates the proposed concept. If this active data selection enables the automatic selection of data for spoof methods that are difficult to deal with in the current detection model and require additional training data, the automatic selection of additional data and continuous training can be triggered and repeated whenever a new deepfake method becomes prevalent. This ensures that the detection system remains effective and up-to-date. Moreover, it is better than blindly adding new data to the `master' training set, since this could lead to an unnecessarily large and imbalanced master set, both in terms of spoofing methods and data distribution.

In this work, we conduct proof-of-concept experiments using the ForgeryNet dataset \cite{he2021forgerynet} as a starter master set and several additional deepfake datasets representing unknown and previously unseen spoofing methods. Our results show that while the ForgeryNet dataset encompasses numerous deepfake techniques, an initial deepfake detector trained solely on this dataset struggles to identify new, unseen spoofing methods accurately. By implementing our proposed continuous training framework, which utilizes automatic active data selection, we demonstrate that the detector's ability can be significantly and efficiently improved compared to one using the starter master set only. We also show that the active selection is better than random selection. 

Section 2 of this paper presents the details of our proposed active data selection strategy for continuous training of deepfake detection models. Experimental conditions and results are provided in Section 3, and we summarize our findings in Section 4. 

\section{Active data selection and continuous training of deepfake detection models}

\subsection{Algorithm}

The algorithm of the proposed method proceeds as follows. First, a starter master dataset $\mathcal{U}_{\text{seed}}$ containing $N$ data from multiple different spoofing methods is prepared, and a robust deepfake detection model is trained from scratch. It is assumed that this starter master dataset covers a certain number of spoofing methods and that the model trained on this dataset will have a certain degree of accuracy. Next, a pool set $\mathcal{U}_{\text{pool}}$ is defined as a set of $M$ data covering many newly emerged spoofing methods. Note that the methods in the pool set are not included in the starter master set. 

Then, the detection model trained from the starter master set is used to make inferences on each data sample $\boldsymbol{x}_{m}$ in the pool set and measure the confidence score $c_m$. Suppose the spoofing method in the pool set is a variant of the ones included in the starter master set. In that case, the detection model is expected to output a reasonable confidence score, while samples generated by completely unseen spoofing methods are expected to have lower confidence scores.

The samples in the pooled set are then sorted based on the confidence scores, and the sample sets with low confidence are selected as a useful data set $\mathcal{I}_{\text{useful}}$ and combined with the seed master set as a new training set $\mathcal{U}_{\text{train}}$. The deepfake detection model is then continuously trained. Specifically, the model is fine-tuned on the combined datasets. Note that this is different from domain adaptation, which only uses new data; in our method, continuous training takes place using a combined set, so information from existing data in $\mathcal{U}_{\text{seed}}$ and additional information in $\mathcal{I}_{\text{useful}}$ are used explicitly. Also note that the labels $\{y_n\}$ of the selected data (and other data in the pool) are assumed to be known. This is reasonable in practice if, for example, the pool data is from a public dataset or generated using APIs.

Finally, the samples added to the new training set are excluded from the pool set, and this process is repeated $K$ times, using a new continuously updated model every time. 

\begin{algorithm}[t]
\caption{Active data selection and continuous training of deepfake detection model.}\label{alg:cap}
\begin{algorithmic}[1]
\Require Starter master set $\mathcal{U}_{\text{seed}}\leftarrow \{\boldsymbol{D}_1, \cdots, \boldsymbol{D}_N\}$ with $N$ data samples
\Require Pool set $\mathcal{U}_{\text{pool}}\leftarrow \{\boldsymbol{D}_{N+1}, \cdots, \boldsymbol{D}_{N+M}\}$ with $M$ data samples

\noindent Note that $\boldsymbol{D}_n=(\boldsymbol{x}_{n}, y_n)$, where $\boldsymbol{x}_{n}$ is an input face image, and $y_n \in\{\text{REAL}, \text{FAKE}\}$ is a label. 
\State $\mathcal{U}_{\text{train}} \leftarrow \mathcal{U}_{\text{seed}}$
\State Model $\leftarrow \text{Training from scratch}(\mathcal{U}_{\text{train}})$
\Repeat 
\For{$\boldsymbol{D}_m \in \mathcal{U}_{\text{pool}}$}
\State $c_m = \mathcal{F}(\boldsymbol{x}_{m}, \text{Model})$ \Comment{Confidence scoring}
\EndFor
\State $\mathcal{I}_{\text{useful}}  \leftarrow \text{argmin-sort}_{m}(\{\cdots, c_{m}, \cdots\})[0:L]$ 
\State $\mathcal{V}_{\text{useful}} \leftarrow \{\boldsymbol{D}_m \in \mathcal{U}_{\text{pool}} | m \in \mathcal{I}_{\text{useful}} \}$ \Comment{Retrieve data with smallest certainty scores}
\State ${\mathcal{U}_{\text{pool}} \leftarrow \mathcal{U}_{\text{pool}} \setminus  \mathcal{V}_{\text{useful}}}$ \Comment{{Remove from pool}}
\State $\mathcal{U}_{\text{train}} \leftarrow \mathcal{U}_{\text{train}} \cup \mathcal{V}_{\text{useful}}$ \Comment{Expand training set}
\State \text{Model} $\leftarrow \text{Continuous training}(\text{Model}, \mathcal{U}_{\text{train}})$ \Comment{Continuous training}
\Until{$K$ iterations are completed}
\end{algorithmic}
\end{algorithm}

\subsection{Negative energy-based confidence score}
\label{sec:method-cm}

The confidence scoring method used in this study is the negative energy-based scoring \cite{NEURIPS2020_f5496252}. Given an input datum $\boldsymbol{x}_{m}$, the model extracts features and transforms them through multiple hidden layers and a softmax output layer. Let the input logits to the softmax layer be $(l_{m,1}, l_{m,2}, \cdots, l_{m,J})$, where $J$ is the number of output classes and $l_{m,j}\in\mathbb{R}, \forall j\in[1,J]$. The certainty score $c_m\in\mathbb{R}$ can be computed by
\begin{equation}
 c_m =  -T\log\sum_{j=1}^{J}\exp(\frac{l_{m,j}}{T}),
\end{equation}
where $T$ is a hyper-parameter dubbed as the softmax temperature. Here, we set $T=1$, following the recipe in \cite{NEURIPS2020_f5496252}. With $c_m$ for each datum in the pool, the useful ones with low $c_m$ are selected (line 6-7 in Algorithm ~\ref{alg:cap}).

The computed $c_m$ is also referred to as a negative energy score in the literature \cite{NEURIPS2020_f5496252}. It links to the energy-based generative model and is utilized for out-of-distribution data detection \cite{NEURIPS2020_f5496252}. The same scoring method has been used in audio deepfake detection models and showed better results than other methods \cite{wangInvestigatingActivelearningbasedTraining2023}. This method is compared with a random selection from a pool set.\footnote{In the experiments described in the next section, strictly speaking, the amount of data per dataset in the pool set has been adjusted and equalized beforehand.}

\section{Experiments}

\subsection{Database}

We conducted experiments using the ForgeryNet dataset \cite{he2021forgerynet} as the starter master set $\mathcal{U}_{\text{seed}}$ and several additional datasets as the pool set $\mathcal{U}_{\text{pool}}$. All datasets used in the experiments are listed in Table~\ref{tab: databases}.

\noindent
\textbf{Starter master set:} We chose the ForgeryNet dataset as the starter master set since it contains 15 different spoofing methods, and hence, it is assumed to be appropriate as the starter master set. We used RetinaFace \cite{9157330} for face detection, and 163,200 facial images were extracted for each of the real and fake classes. Then, the bounding box was enlarged by a factor of 1.3. The extracted face image was resized to 384$\times$384 using bicubic interpolation.

\noindent
\textbf{Pool set:} The pool set consists of multiple databases. Considering the use of different or newer spoofing methods than those included in ForgeryNet, face images from the FF++ \cite{rossler2019faceforensics++}, Google DFD \cite{googledfd}, YouTube DF \cite{kukanov2020cost}, KoDF \cite{kwon2021kodf}, and Stable Diffusion 2.1 \cite{rombach2022high} datasets were used as data in the fake class of the pool set. As for the data in the real class, in addition to the real parts of the aforementioned databases above, VoxCeleb \cite{chung2018voxceleb2} and FFHQ \cite{karras2019style} were also used. The same face extraction and pre-processing as in the starter master set were applied to construct a pool set of 40,000 real and fake face images from each dataset.

\noindent
\textbf{Validation and test sets:} The validation and test sets include 1,000 face images selected from each class in each dataset.

\begin{table}[t!]
\centering
\caption{Dataset design used in this paper.}
\label{tab: databases}
\begin{tabular}{llllll} 
\hline
\textbf{Database} & \textbf{Type} & \textbf{Initial} & \textbf{AL Pool} & \textbf{Val.} & \textbf{Test} \\ 
\hline
\textbf{\textit{Starter master set}}\\
ForgeryNet \cite{he2021forgerynet} & Real & 163,200 &  & 1,000 & 1,000 \\
ForgeryNet \cite{he2021forgerynet} & Fake & 163,200 &  & 1,000 & 1,000 \\ 
\hdashline
\textbf{\textit{Pool set}}\\
FF++ \cite{rossler2019faceforensics++} & Real &  & 40,000 & 1,000 & 1,000 \\
FF++ (5 types) \cite{rossler2019faceforensics++} & Fake &  & 40,000 & 1,000 & 1,000 \\
Google DFD \cite{googledfd} & Real &  & 40,000 & 1,000 & 1,000 \\
Google DFD \cite{googledfd} & Fake &  & 40,000 & 1,000 & 1,000 \\
VoxCeleb \cite{chung2018voxceleb2} & Real &  & 40,000 & 1,000 & 1,000 \\
YouTube DF \cite{kukanov2020cost} & Fake &  & 40,000 & 1,000 & 1,000 \\
KoDF \cite{kwon2021kodf} & Real &  & 40,000 & 1,000 & 1,000 \\
KoDF \cite{kwon2021kodf} & Fake &  & 40,000 & 1,000 & 1,000 \\
FFHQ \cite{karras2019style} & Real &  & 40,000 & 1,000 & 1,000 \\
Stable Diffusion 2.1 \cite{rombach2022high} & Fake &  & 40,000 & 1,000 & 1,000 \\ 
\hline
\end{tabular}
\end{table}

\subsection{Deepfake detection system used in experiments}

Our detector uses the EfficientNet V2-M architecture \cite{pmlr-v139-tan21a} pre-trained by ImageNet21k \cite{deng2009imagenet} as a backbone, with a head layer that makes binary predictions of real and fake. The model was trained using AdamW with a learning rate of $5\times10^{-4}$. The batch size was set to 128. During model training, data augmentation was carried out in a similar way to DeepfakeBench, a benchmark platform for deepfake detection \cite{yan2023deepfakebench}. The best checkpoint among 100 training epochs was selected on the basis of the loss of the validation set. 

\subsection{Systems to be compared}

The experiments included the following three systems:
\begin{itemize}
\item \texttt{Base} is the baseline system trained using the starter master set without active data selection or continuous training.
\item \texttt{AL\_negE} is based on \texttt{Base} but further trained using the active data selection and continuous training utilizing the pool set. The confidence score is measured using the negative energy score explained in Section~\ref{sec:method-cm}.
\item \texttt{AL\_random} is a reference system configured in the same way as \texttt{AL\_negE} except that the training data is randomly selected from the pool set. 
\end{itemize}

The \texttt{AL\_negE} and \texttt{AL\_random} systems used the same optimizer and learning rate as \texttt{Base} through the continuous training loops. The number of samples selected from the pool in a single continuous training iteration was 10,000 (i.e., $L=10,000$ in Algorithm ~\ref{alg:cap}). The fine-tuning was conducted for three epochs per continuous training iteration (line 10 in Algorithm ~\ref{alg:cap}).

\subsection{Results}

\noindent
\textbf{Evaluation metric} 

System performance is reported via equal error rates (EERs)~\cite{iso2017iso}. The EER corresponds to the percentage of errors when the decision threshold of the system is set such that the rates of false acceptance (i.e., a fake image being classified as \emph{real}) and false rejection (i.e., a real image being classified as \emph{fake}) are equal. A lower EER indicates a better performance. 

Note that we intentionally choose EER, even though it requires an oracle decision threshold (i.e., by assuming that we know the labels of the test set). EER gauges the discriminative power of the detector with being muddled up with the calibration~\cite{castro2007forensic}. Other evaluation metrics are left for future work.

\noindent
\textbf{Results of the \texttt{Base} model} 

\begin{figure}[t!]
\centering
\includegraphics[trim=0 0 0 0, clip, width=0.75\columnwidth]{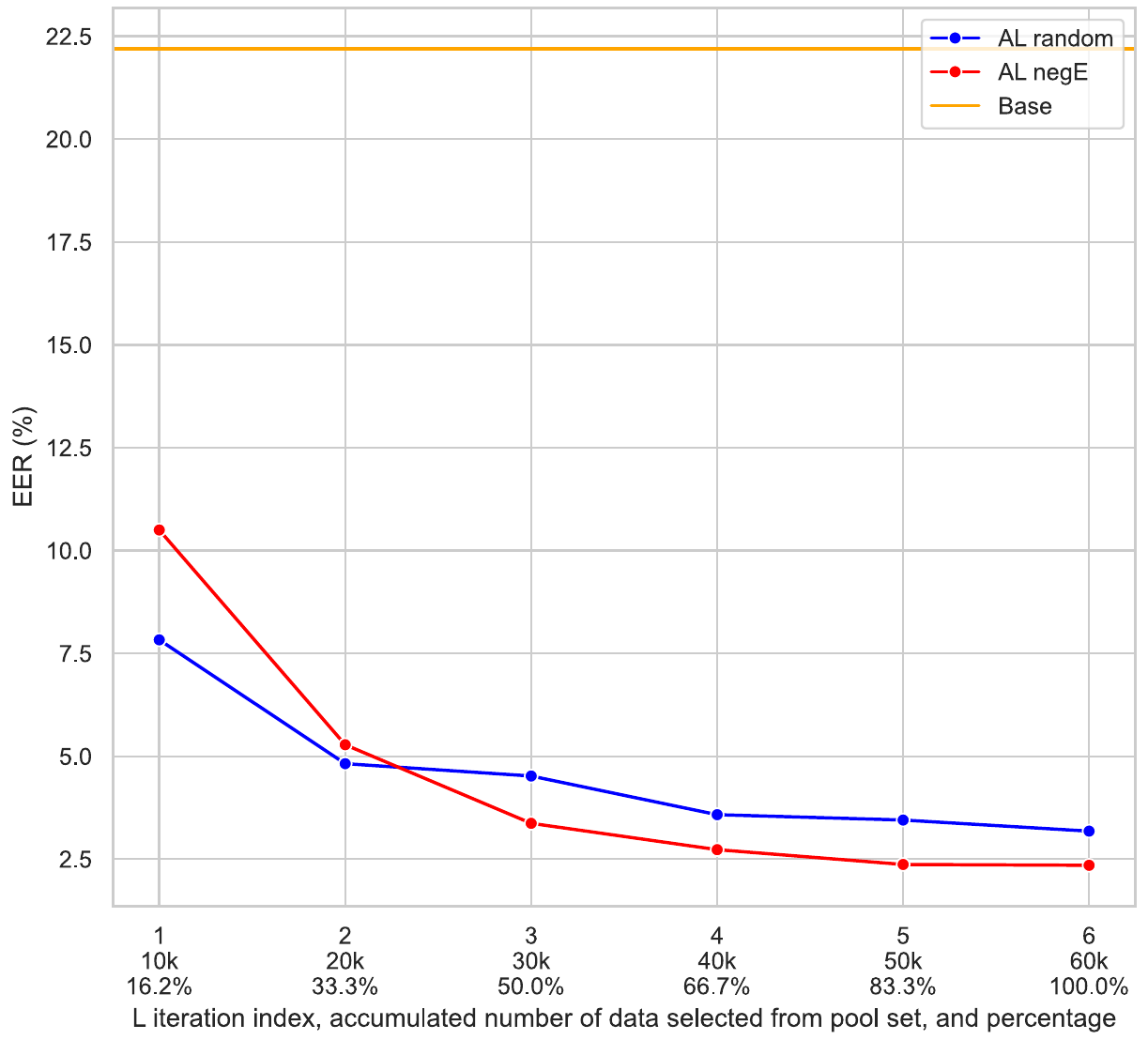}
\caption{EERs (\%) on the evaluation set across different continuous training iterations. Numbers along the horizontal axis are the iteration index, number of data samples selected from the pool set, and its percentage.} 
\label{fig:exp_error}
\end{figure}

We first look at the \texttt{Base} model trained on the ForgeryNet dataset, which contains 15 different spoofing methods, and evaluate it on the ForgeryNet-only validation set. In this case, the EER was 2.1\% (not shown in Fig. \ref{fig:exp_error}). This result indicates that the model was trained appropriately. However, looking at the result in Fig. \ref{fig:exp_error}, we can see that the \texttt{Base} model, as expected, fails to adequately detect spoofing methods in the test set containing multiple sources (shown in Table~\ref{tab: databases}), resulting in a very high EER of 22.5\%.

\noindent
\textbf{Results of the \texttt{AL\_negE} and \texttt{AL\_random} models} 

Next, we focus on the results of active data selection or random selection and update the model through continuous training. As shown in Fig. \ref{fig:exp_error}, first, the detection performance can be significantly improved by adding 10,000 selected images in each iteration and by performing continuous training of the model. Also, in the first and second iterations, random selection gave better results, but in subsequent iterations, the proposed active data selection was able to select better data from the pool set and reduce the EER to just under 2.5\% even though \textit{only 15\% of the data of the pool set was used}.\footnote{This also significantly changes the time needed for model training. If we were to add all data in the pool set to the master set and train the model, it would take 2798 sec/epoch on our server, whereas if only the data selected by the proposed method is added to the master set, it takes only 1303 sec/epoch.}

The random set also showed improvement, partly because the data sets in the pool set were processed to be equal in quantity beforehand, which reduces bias to a large extent. If there is a bias in the pool set, more data from the spoofing methods predominant in the pool set will be selected, and hence, the improvement would not be as good as in Fig~\ref{fig:exp_error}.

\subsection{Analysis}

\begin{figure}[t!]
\centering
\includegraphics[width=1.0\columnwidth]{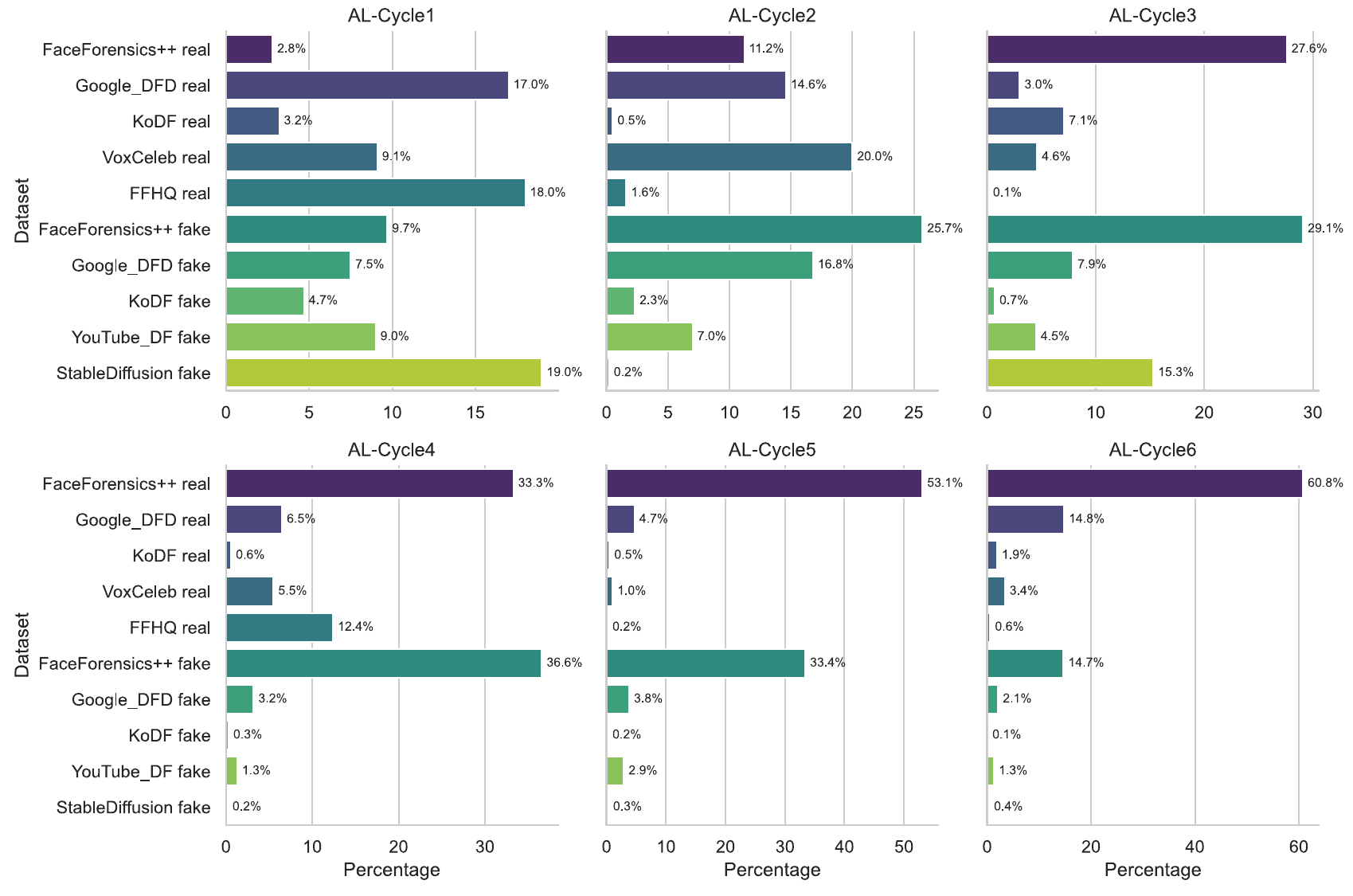}
\caption{Percentage indicating from which dataset the image was selected in each iteration.} 
\label{fig:exp_data}
\end{figure}

Figure~\ref{fig:exp_data} shows from which datasets the images were selected during the proposed active data selection. Interestingly, images from the newer method, such as Stable Diffusion, were selected in the first half of the iteration, while images from the third iteration were selected from the relatively old database, FF++. It can also be seen that not many images were selected from the KoDF dataset. Presumably, the spoofing methods included in this dataset can be detected by a model trained using the ForgeryNet dataset. 

\section{Conclusions}

In this paper, we proposed a method for automatically and actively selecting the small amount of additional data required for continuous training of a deepfake detection model in situations where the model is regularly updated. The policy is to use the confidence level of the deepfake detection model itself and automatically select from a redundant pool set the data needed to improve the performance of the model. Experimental results showed that the deepfake detection model continuously trained with a small amount of additional data added to the starter master set significantly and efficiently improved the detection performance, reducing the EER to 2.5\% with only 15\% of the pool set of data.

Our future work will include looking at different data selection metrics and testing system performance on data of domains different from those in the pool set. Considering the fact that deepfake is being actively created on the Internet, we also plan to test the system performance using deepfake in the wild.

\section*{Acknowledgements}

This study is partially supported by JST CREST Grants (JPMJCR18A6, JPMJCR20D3), JST AIP Acceleration Research (JPMJCR24U3), and MEXT KAKENHI Grants (24H00732). This research was also partially supported by the project for the development and demonstration of countermeasures against disinformation and misinformation on the Internet with the Ministry of Internal Affairs and Communications of Japan. This study was carried out using the TSUBAME4.0 supercomputer at Tokyo Institute of Technology.
\\
\\
\\
\\
\bibliography{main}

\begin{thebibliography}{KLA19}

\bibitem[Bi17]{iso2017iso}
Biometrics, ISO/IEC JTC1~SC37: , ISO/IEC 2382-37: 2017 Information Technology-Vocabulary-Part 37: Biometrics, 2017.

\bibitem[Ca07]{castro2007forensic}
Castro, Daniel~Ramos: Forensic Evaluation of the Evidence Using Automatic Speaker Recognition Systems.
\newblock Dissertation, Universidad aut{\'o}noma de Madrid, 2007.

\bibitem[CNZ18]{chung2018voxceleb2}
Chung, Joon~Son; Nagrani, Arsha; Zisserman, Andrew: {VoxCeleb2}: Deep Speaker Recognition.
\newblock In: Proc. {INTERSPEECH}.
\newblock S. 1086--1090, 2018.

\bibitem[Da24]{das2024limited}
Das, Srijan; Jain, Tanmay; Reilly, Dominick; Balaji, Pranav; Karmakar, Soumyajit; Marjit, Shyam; Li, Xiang; Das, Abhijit; Ryoo, Michael~S: Limited Data, Unlimited Potential: A Study on ViTs Augmented by Masked Autoencoders.
\newblock In: Proc. WACV.
\newblock S. 6878--6888, 2024.

\bibitem[De09]{deng2009imagenet}
Deng, Jia; Dong, Wei; Socher, Richard; Li, Li-Jia; Li, Kai; Fei-Fei, Li: {ImageNet}: A large-scale hierarchical image database.
\newblock In: Proc. CVPR.
\newblock IEEE, S. 248--255, 2009.

\bibitem[De20]{9157330}
Deng, Jiankang; Guo, Jia; Ververas, Evangelos; Kotsia, Irene; Zafeiriou, Stefanos: RetinaFace: Single-Shot Multi-Level Face Localisation in the Wild.
\newblock In: Proc. CVPR.
\newblock S. 5202--5211, 2020.

\bibitem[DG19]{googledfd}
Dufour, Nick; Gully, Andrew: , Contributing Data to Deepfake Detection Research.
\newblock \url{https://ai.googleblog.com/2019/09/contributing-data-to-deepfake-detection.html}, 9 2019.

\bibitem[Do23]{dong2023contrastive}
Dong, Fengkai; Zou, Xiaoqiang; Wang, Jiahui; Liu, Xiyao: Contrastive learning-based general Deepfake detection with multi-scale RGB frequency clues.
\newblock Journal of King Saud University-Computer and Information Sciences, 35(4):90--99, 2023.

\bibitem[He21]{he2021forgerynet}
He, Yinan; Gan, Bei; Chen, Siyu; Zhou, Yichun; Yin, Guojun; Song, Luchuan; Sheng, Lu; Shao, Jing; Liu, Ziwei: {ForgeryNet}: A versatile benchmark for comprehensive forgery analysis.
\newblock In: Proc. CVPR.
\newblock S. 4360--4369, 2021.

\bibitem[KLA19]{karras2019style}
Karras, Tero; Laine, Samuli; Aila, Timo: A style-based generator architecture for generative adversarial networks.
\newblock In: Proc. CVPR.
\newblock S. 4401--4410, 2019.

\bibitem[Ku20]{kukanov2020cost}
Kukanov, Ivan; Karttunen, Janne; Sillanp{\"a}{\"a}, Hannu; Hautam{\"a}ki, Ville: Cost sensitive optimization of deepfake detector.
\newblock In: Proc. APSIPA ASC.
\newblock IEEE, S. 1300--1303, 2020.

\bibitem[Kw21]{kwon2021kodf}
Kwon, Patrick; You, Jaeseong; Nam, Gyuhyeon; Park, Sungwoo; Chae, Gyeongsu: Kodf: A large-scale korean deepfake detection dataset.
\newblock In: Proc. ICCV.
\newblock S. 10744--10753, 2021.

\bibitem[La23]{larue2023seeable}
Larue, Nicolas; Vu, Ngoc-Son; Struc, Vitomir; Peer, Peter; Christophides, Vassilis: SeeABLE: Soft Discrepancies and Bounded Contrastive Learning for Exposing Deepfakes.
\newblock In: Proc. ICCV.
\newblock S. 21011--21021, 2023.

\bibitem[Li20]{NEURIPS2020_f5496252}
Liu, Weitang; Wang, Xiaoyun; Owens, John; Li, Yixuan: {Energy-based Out-of-distribution Detection}.
\newblock In: Proc. NIPS.
\newblock Jgg.~33, S. 21464--21475, 2020.

\bibitem[Li23]{liu2024forgery}
Liu, Huan; Tan, Zichang; Tan, Chuangchuang; Wei, Yunchao; Zhao, Yao; Wang, Jingdong: Forgery-aware Adaptive Transformer for Generalizable Synthetic Image Detection.
\newblock In: Proc. CVPR.
\newblock 2023.

\bibitem[Ma21]{ma2021continual}
Ma, Haoxin; Yi, Jiangyan; Tao, Jianhua; Bai, Ye; Tian, Zhengkun; Wang, Chenglong: Continual learning for fake audio detection.
\newblock In: Proc. INTERSPEECH.
\newblock 2021.

\bibitem[Ro19]{rossler2019faceforensics++}
Rossler, Andreas; Cozzolino, Davide; Verdoliva, Luisa; Riess, Christian; Thies, Justus; Nie{\ss}ner, Matthias: {FaceForensics++}: Learning to detect manipulated facial images.
\newblock In: Proc. ICCV.
\newblock S. 1--11, 2019.

\bibitem[Ro22]{rombach2022high}
Rombach, Robin; Blattmann, Andreas; Lorenz, Dominik; Esser, Patrick; Ommer, Bj{\"o}rn: High-resolution image synthesis with latent diffusion models.
\newblock In: Proc. CVPR.
\newblock S. 10684--10695, 2022.

\bibitem[SLL24]{song2024towards}
Song, Wentang; Lin, Yuzhen; Li, Bin: Towards Generic Deepfake Detection with Dynamic Curriculum.
\newblock In: Proc. ICASSP.
\newblock IEEE, S. 4500--4504, 2024.

\bibitem[TL21]{pmlr-v139-tan21a}
Tan, Mingxing; Le, Quoc: EfficientNetV2: Smaller Models and Faster Training.
\newblock In (Meila, Marina; Zhang, Tong, Hrsg.): Proceedings of the 38th International Conference on Machine Learning.
\newblock Jgg. 139 in Proc. Machine Learning Research. PMLR, S. 10096--10106, 18--24 Jul 2021.

\bibitem[WY23]{wangInvestigatingActivelearningbasedTraining2023}
Wang, Xin; Yamagishi, Junichi: Investigating {{Active-learning-based Training Data Selection}} for {{Speech Spoofing Countermeasure}}.
\newblock In: Proc. {{SLT}}.
\newblock S. 585--592, 2023.

\bibitem[Ya23]{yan2023deepfakebench}
Yan, Zhiyuan; Zhang, Yong; Yuan, Xinhang; Lyu, Siwei; Wu, Baoyuan: DeepfakeBench: A Comprehensive Benchmark of Deepfake Detection.
\newblock In: Proc. NeurIPS Datasets and Benchmarks Track.
\newblock 2023.

\bibitem[Ya24]{yan2024deepfakebench}
Yan, Zhiyuan; Zhang, Yong; Yuan, Xinhang; Lyu, Siwei; Wu, Baoyuan: {DeepfakeBench}: A Comprehensive Benchmark of Deepfake Detection.
\newblock Advances in Neural Information Processing Systems, 36, 2024.

\end{thebibliography}

\end{document}